\documentclass[
]{ceurart}

\usepackage{graphicx}
\usepackage{subcaption}
\usepackage[font=small,labelfont=bf]{caption}
\usepackage{float}

\usepackage{placeins}
\usepackage{listings}
\begin{document}

\copyrightyear{2025}
\copyrightclause{Copyright for this paper by its authors.
  Use permitted under Creative Commons License Attribution 4.0
  International (CC BY 4.0).}

\conference{AEQUITAS 2025: Workshop on Fairness and Bias in AI | co-located with ECAI 2025, Bologna, Italy}

\title{Visual Model Selection using Feature Importance Clusters in Fairness-Performance  Similarity Optimized Space}

\author[1]{Sofoklis Kitharidis}[%
orcid=0009-0005-8404-0724,
email=s.kitharidis@liacs.leidenuniv.nl,
]
\address[1]{Leiden Institute of Advanced Computer Science (LIACS), Leiden University, Einsteinweg 55, 2333~CC Leiden, The Netherlands}
\address[2]{Netherlands Organization for Applied Scientific Research (TNO), Anna van Buerenplein 1, 2595~DA The Hague, The Netherlands}

\author[1,2]{Cor J. Veenman}[%
orcid=0000-0002-2645-1198,
email=cor.veenman@tno.nl,
]

\author[1]{Thomas B\"ack}[%
orcid=0000-0001-6768-1478,
email=t.h.w.baeck@liacs.leidenuniv.nl,
]

\author[1]{Niki van Stein}[%
orcid=0000-0002-0013-7969,
email=n.van.stein@liacs.leidenuniv.nl,
]

\begin{abstract}
In the context of algorithmic decision-making, fair machine learning methods often yield multiple models that balance predictive fairness and performance in varying degrees. This diversity introduces a challenge for stakeholders who must select a model that aligns with their specific requirements and values. To address this, we propose an interactive framework that assists in navigating and interpreting the trade-offs across a portfolio of models. Our approach leverages weakly supervised metric learning to learn a Mahalanobis distance that reflects similarity in fairness and performance outcomes, effectively structuring the feature importance space of the models according to stakeholder-relevant criteria. We then apply clustering technique (k-means) to group models based on their transformed representations of feature importances, allowing users to explore clusters of models with similar predictive behaviors and fairness characteristics. This facilitates informed decision-making by helping users understand how models differ not only in their fairness-performance balance but also in the features that drive their predictions.
\end{abstract}


\maketitle

\section{Introduction}
Ensuring that machine learning systems not only excel in predictive accuracy but also uphold equitable treatment is crucial, especially in high-stakes domains where decisions can profoundly impact individuals’ lives. Over the past decade, algorithmic fairness has emerged as a critical consideration, aiming to ensure models do not disadvantage protected groups (e.g. by gender or race) \cite{survey_fairness}. Numerous quantitative fairness definitions have been proposed, such as demographic parity, equalized odds, and predictive parity, but these criteria often cannot all be satisfied simultaneously \cite{impossib_theorem}. Developers thus face fundamental trade-offs between model fairness and performance: improving a fairness metric usually degrades accuracy or violate another fairness notion. As a result, achieving “fair” machine learning is inherently a multi-objective problem that requires context-dependent value judgments \cite{impossib_theorem}. This challenge is intensified by the complexity of real-world data and biases, making fairness an active and complex area of research.

Organizations seeking to implement decision-making algorithms frequently encounter significant technical challenges. Bridging fairness research to practice remains difficult, as existing fairness mitigation algorithms, ranging from pre-processing data transformations to in-processing model constraints and post-processing outcome adjustments \cite{survey_fairness}, have not fully translated into user-friendly tools for organizations. Therefore, there is an urgent need for human-centered frameworks to support fair decision-making within applied contexts. Stakeholders such as domain experts, policymakers, or model auditors need to understand the trade-offs involved in selecting one model over another. For instance, different classifiers or hyperparameter settings can yield comparable overall accuracy yet produce significantly varied fairness outcomes \cite{intentionalfairness}. One model might maximize predictive accuracy while exhibiting higher bias against a minority group, whereas an alternative model may sacrifice a small amount of accuracy for a more equitable distribution of errors. Choosing among these models requires not only technical evaluation but also alignment with broader societal and stakeholder values. However, without dedicated support, it is challenging for decision-makers to fully understand these trade-offs, particularly when multiple fairness metrics and model behaviors must be compared simultaneously.

In many cases, decision-makers are presented with a collection of models that embody different fairness–performance trade-offs, rather than a single “best” solution. Systematically exploring and comparing a large set of models is non-trivial since it can be overwhelming to evaluate dozens of models across multiple fairness and performance metrics, particularly when each model may rely on different features in subtle ways. Existing visualization tools, such as Fairlearn’s dashboard \cite{fairlearn} and IBM AI Fairness 360 \cite{ai360}, typically focus on individual models or isolated fairness-performance trade-offs, limiting the stakeholders' capacity to understand the complete landscape of available models simultaneously. 

In contrast, our framework arranges models in a two-dimensional fairness–performance space and overlays clusters derived from their transformed feature-importance profiles. By grouping models whose decision logic reflects similar feature attributions, we surface archetypes whose behavior aligns with particular domain hypotheses or known causal relationships. Stakeholders can therefore choose not only based on a cluster’s position in the fairness–performance space, but also because its characteristic feature-importance signature resonates with organizational policies or domain expertise.

\section{Related Works}
In this section, we review core algorithmic approaches for enforcing fairness in machine learning, interpretability techniques that shed light on model behavior, and interactive visualization tools that support human‐centered model comparison.

\paragraph{Fairness in Machine Learning.}\mbox{}\\
Ensuring fairness in predictive models has been the focus of extensive research, yielding various definitions and mitigation strategies \cite{survey_fairness}. Broadly, fairness interventions are categorized as pre-processing (altering training data), in-processing (altering the optimization algorithm), or post-processing (altering model outputs) \cite{survey_fairness}.  Our work concerns \textbf{in-processing} techniques that directly build models capable of achieving different fairness-performance trade-offs. One line of research adds fairness constraints or objectives into model training. For example, Agarwal et al. (2018) \cite{flogistic} formulate fairness-constrained classification as a series of cost-sensitive learning tasks, finding a model with minimal error subject to a fairness constraint. This reductions-based approach can enforce criteria like demographic parity or equalized odds by adjusting weights on training examples, and it has become a general blueprint implemented in toolkits (e.g. Microsoft’s Fairlearn library implements this strategy \cite{fairlearn}). Another approach is to design custom learning algorithms that inherently balance accuracy and fairness. Barata et al. (2021) \cite{fairtreeclassifier} introduce a splitting criterion for decision trees that combines ROC AUC with a fairness measure (strong demographic parity) during each split. By optimizing a trade-off of fairness and performance at training time, it produces decision trees that are interpretable and explicitly designed to achieve specific fairness-performance trade-offs. \cite{fairtreeclassifier}. In practice, data scientists must often adjust a hyperparameter (such as the fairness penalty strength or a target constraint value) to get a model that achieves an acceptable balance. This tuning yields multiple candidate models along a continuum from highest accuracy to highest fairness, rather than a single optimal solution.

\paragraph{Interpretability and Fairness Analysis.}\mbox{}\\
Common \textbf{post-hoc} interpretability techniques include feature importance measures and instance-level explanations. \textbf{SHAP} values \cite{shap} have become a popular choice for explaining complex models because they provide consistent and theoretically grounded attributions for each feature’s influence on a prediction. Specifically, \textbf{SHAP} uses cooperative game theory to calculate the marginal contributions of each feature by considering all possible subsets of features. In the context of fairness, \textbf{SHAP} and related methods have been used to diagnose which features might be contributing to bias. For example, Cabrera et al. (2019) \cite{fairvis} used subgroup analysis and feature attributions to discover that certain features caused disproportionate errors for specific demographic subgroups. Another common approach to feature importance is \textbf{permutation importance} \cite{permutation}, an intuitive technique where a feature’s values are randomly shuffled to see how much model error increases. This provides a global ranking of features by their influence on model performance.

\paragraph{Interactive and Human-Centered Tools for Model Selection.}\mbox{}\\
The \textbf{What-If Tool (WIT)} from Google’s PAIR initiative \cite{wit} allows users to visualize classification metrics, manipulate test inputs, and compare outcomes across different models in a dashboard interface. Other tools like \textbf{Fairlearn’s} dashboard (by Microsoft) \cite{fairlearn} and \textbf{IBM AI Fairness 360} \cite{ai360} provide visualizations of fairness metrics, but these generally focus on assessing one model at a time or adjusting single-model thresholds, such as classification decision boundaries. In the research community, visualization systems such as \textbf{FairSight} \cite{fairsight} and \textbf{FairVis} \cite{fairvis} have explored ways to involve end-users in fairness auditing. \textbf{FairSight} provides a comprehensive visual analytics workflow to understand, diagnose, and mitigate biases in ranking decisions. \textbf{FairVis}, on the other hand, helps users discover intersectional biases by comparing subgroup performance within a single model. However, these existing methods often lack the capability to simultaneously visualize multiple models comprehensively, limiting stakeholders’ ability to effectively compare diverse fairness-performance trade-offs. Our approach addresses this gap by providing stakeholders with a visualization framework that facilitates comprehensive comparison across an entire set of candidate models and allows focused exploration of specific model aligned with stakeholder priorities.

\section{Clustering Fair Learners} \label{methodology}

Our proposed framework is designed to help stakeholders navigate and interpret fairness-performance trade-offs within a large set of predictive models. The methodology integrates three core components: (1) weakly supervised metric learning based on fairness-performance proximity, (2) a feature importance transformation using this metric, and (3) clustering models based on transformed feature importance profiles. Below we provide a detailed description of each stage.

We begin by assuming the availability of a diverse set of models produced by fairness-aware learning methods with varying fairness‐penalty settings~$\theta$ (e.g., classifiers trained with fairness constraints, or hyperparameter tuning). Each model ~$m$ is characterized by:

\begin{itemize}
\item A predictive performance metric \emph{perf} (e.g., accuracy, AUC),
\item A fairness metric \emph{fair} (e.g., demographic parity or equalized odds),
\item A vector of \emph{feature‐importance} scores $\mathbf{x}_m \in \mathbb{R}^P$ (e.g., SHAP \cite{shap} or permutation importances \cite{permutation}).
\end{itemize}

Since, we need a way to characterize models in terms of their inferencing behavior; feature-importance values serve as a proxy for this, offering interpretable profiles of how each feature influences predictions and empowering decision-makers to justify their model selection based on feature usage patterns.

\subsection{Weakly Supervised Metric Learning}
Clustering raw feature-importance vectors directly using a Euclidean distance metric is inadequate, as it assumes all dimensions are equally informative and comparable in scale. In practice, some feature‐importance dimensions capture critical distinctions in model behaviour, while others contribute only noise; treating them uniformly can obscure meaningful structure. Moreover, disparities in variable scale can dominate distance computations, yielding groupings that reflect arbitrary scale differences rather than stakeholder‐relevant similarities in fairness–performance trade-offs. Consequently, we propose structuring the model space through \textbf{weakly supervised metric learning}, making the stakeholders' viewpoint on model similarity clearer and more explicit.

We consider the fairness-performance space as representative for the perceived nearness of models to the decision-maker. Therefore, by learning a suitable transformation of feature-importance vectors, we produce an embedding where proximity directly corresponds to similarity in fairness–performance trade-offs. Consequently, models that achieve similar balances of accuracy and fairness are embedded close together, facilitating meaningful grouping and interpretability. Conversely, models with substantially different positions in the fairness–performance space are pushed apart in the transformed space, while those sharing a similar trade‐off remain clustered. Specifically, we employ \textbf{Information-Theoretic Metric Learning (ITML)} \cite{itml} to learn a Mahalanobis distance. ITML’s information‐theoretic formulation yields a positive‐definite, well‐conditioned metric under flexible constraints, avoiding trivial identity‐matrix solutions and slow convergence issues common to alternative methods \cite{metric_learning_dis}.

\paragraph{Pairwise Constraints from Fairness-Performance Space}\mbox{}\\
To guide metric learning effectively, we establish weak supervision through pairwise constraints derived from the joint fairness-performance space. Specifically, we first calculate pairwise Euclidean distances between all models based on their positions in the fairness-performance space. We empower stakeholders to explicitly define thresholds that characterize which models are considered similar or dissimilar in their specific decision context. In our implementation, similarity and dissimilarity thresholds are explicitly set (e.g., 0.05 and 0.2, respectively) to directly reflect stakeholder preferences regarding model similarity in terms of fairness and performance trade-offs.

Pairs of models whose transformed distances are below the similarity threshold (e.g., 0.05) are labeled as similar, whereas pairs exceeding the dissimilarity threshold (e.g., 0.2) are labeled dissimilar. Given potential imbalances between the counts of similar and dissimilar pairs, we enforce balance by subsampling from the larger set, resulting in equal representation of both constraint types.

\paragraph{Metric Learning via ITML}\mbox{}\\
Leveraging the established pairwise constraints, we employ ITML to learn a Mahalanobis distance metric represented by the positive-definite matrix \emph{M}. ITML optimizes a LogDet-regularized objective, ensuring robustness even with noisy or sparse constraint data. Formally, the Mahalanobis distance between two feature-importance vectors $\mathbf{x}_i$ and $\mathbf{x}_j$ under this learned metric \emph{M} is computed as:

\[
d_M(\mathbf{x}_i, \mathbf{x}_j)
\;=\;
\sqrt{\bigl(\mathbf{x}_i - \mathbf{x}_j\bigr)^\mathsf{T}\,M\,\bigl(\mathbf{x}_i - \mathbf{x}_j\bigr)},
\]

where \emph{M} is the learned positive-definite Mahalanobis matrix \cite{Mahalanobis1936}. Intuitively, ITML learns a tailored distance measure to align better with user-defined similarity constraints, effectively emphasizing or de-emphasizing certain feature differences based on the provided weak supervision. Unlike standard Euclidean distance, which treats all feature differences equally, ITML adjusts the scale and correlation between features, ensuring models that stakeholders perceive as similar in fairness–performance trade-offs appear closer, while models with contrasting trade-offs are pushed further apart. This targeted adjustment facilitates more meaningful and interpretable clustering outcomes. In implementation, ITML (metric-learn) is trained on balanced sets of similar/dissimilar pair constraints; zero-distance pairs in the original feature-importance space are excluded, and an identity prior with \texttt{max\_iter}=600 is employed.

\subsection{Clustering in the Learned Space}
Using the learned metric to transform the original feature importance vectors, we systematically organize models into clusters that reflect coherent and interpretable fairness-performance profiles. In our experiments, we employ \textbf{k-means clustering} to partition the transformed model space. Because the ITML transformation produces a Mahalanobis space in which similarity constraints tend to form roughly spherical groups \cite{itml}, k-means is particularly well-suited to capture these cluster shapes efficiently and interpretably. Specifically, it is used with
\emph{k-means++} initialization, \texttt{n\_init}=10 restarts, and a fixed
\texttt{random\_state}=42; for each $k$ we retain the run with the lowest
within-cluster sum of squares (inertia). While alternative techniques (e.g., hierarchical clustering, DBSCAN, or Gaussian mixture models) could be explored, k-means aligns directly with our spherical-cluster assumption. 

Selecting the optimal number of clusters is essential but inherently challenging, given the complexity and variability of real-world data distributions. Extensive comparative studies emphasize that no single internal cluster validation index (CVI) consistently outperforms \cite{Gagolewski_2021}. Each CVI inherently biases towards specific cluster characteristics—some favor compact, spherical clusters, whereas others better handle elongated or irregularly shaped clusters. As a result, relying on a single CVI often yields conflicting recommendations for the optimal number of clusters \cite{Gagolewski_2021}.

\paragraph{Composite Validation for Optimal $k$ Selection}\mbox{}\\
To address this challenge robustly, we implement a composite validation strategy leveraging multiple internal CVIs to determine the optimal number of clusters. Specifically, for each candidate number of clusters ($k\in\{3,\dots,20\}$), we evaluate clustering solutions using the following complementary indices:

\begin{itemize}
\item \textbf{Silhouette Score} \cite{Silhouettes}, which for each point measures how much closer it is to points in its own cluster than to points in the nearest other cluster, and then averages over all points, capturing the \emph{average cohesion versus separation}.
\item \textbf{Calinski–Harabasz Index} \cite{Caliński01011974}, defined as the ratio of between-cluster dispersion to within-cluster dispersion, reflecting the \emph{global compactness and separation} of the partition.
\item \textbf{Davies–Bouldin Index} \cite{Davies}, which computes for each cluster the maximum ratio of the sum of its intra-cluster scatter to the inter-cluster separation with its most similar cluster, then averages these maxima, quantifying the \emph{average worst-case cluster similarity}.
\item \textbf{Dunn Index} \cite{Dunn}, taking the minimum inter-cluster distance divided by the maximum intra-cluster diameter, emphasizing the \emph{worst-case separation} relative to cluster tightness.
\end{itemize}

Each metric thus highlights a unique perspective: Silhouette focuses on point-level cohesion, Calinski–Harabasz on overall dispersion ratios, Davies–Bouldin on penalizing clusters that are too alike, and Dunn on guarding against poorly separated clusters. To integrate their strengths, we standardize (z-score) each metric across all $k$, invert Davies–Bouldin so higher is better, and compute

\begin{equation}
  \text{composite}(k)
  = z_{\mathrm{Sil}}(k) + z_{\mathrm{CH}}(k) + z_{\mathrm{DB}}(k) + z_{\mathrm{Dunn}}(k)
  \label{eq:composite}
\end{equation}
\begin{equation}
  k^* = \arg\max_k \text{composite}(k).
  \label{eq:kstar}
\end{equation}

\section{Experiments}

In this section, we empirically validate our proposed interactive framework for clustering and interpreting fairness–performance trade-offs across predictive models and datasets. Specifically, we investigate how effectively our weakly supervised metric learning approach organizes models into meaningful groups, assess the interpretability of resulting clusters, and examine the robustness of our methodology across different datasets and fairness-aware learning methods.

\subsection{Setup} \label{exper_setup}

\paragraph{Datasets}\mbox{}\\
We evaluate our proposed method on UCI Machine Learning Repository datasets \cite{UCI} that exemplify fairness-sensitive decision-making tasks. First, the \textbf{Adult dataset} which is used to predict whether an individual’s income exceeds \$50K per year; the sensitive attributes considered are \textit{race}, \textit{gender}, and \textit{age}. Second, the \textbf{Bank Marketing dataset} is employed to predict whether a client subscribes to a term deposit following a marketing campaign, with \textit{age} treated as the sensitive attribute. Both datasets include a mixture of numerical and categorical variables.

\paragraph{Fairness-Aware Methods}\mbox{}\\
We evaluate our clustering framework using two representative fairness-aware learning methods, each parameterized by a hyperparameter that controls the fairness-performance trade-off.

The \textbf{FairTree Classifier (FTC)}~\cite{fairtreeclassifier} is a decision-tree algorithm that optimizes splits based on a compound criterion (\emph{SCAFF}) combining predictive performance (ROC AUC) and fairness with respect to strong demographic parity (SDP). FTC introduces an orthogonality parameter $\Theta \in [0,1]$, defined as:
\[
\text{SCAFF} = (1 - \Theta) \cdot \text{ROC-AUC} + \Theta \cdot \text{SDP},
\]
where $\Theta=0$ corresponds to maximizing pure predictive accuracy and $\Theta=1$ enforces maximal fairness by prioritizing sensitive-attribute parity. SDP is computed by, for each sensitive attribute (for example race, gender, age) and for each of its categories, measuring a one-versus-rest disparity score and then taking the worst-case (minimum) parity across all attributes and groups.

The \textbf{Fair Logistic Regression (FLR)} method~\cite{flogistic} formulates fair classification as a constrained optimization problem, solvable via Lagrangian duality. The method integrates fairness constraints through Lagrange multipliers $\lambda$, which are regularized by an $\ell_1$-norm bound $B \in [0,\infty)$:
\[
\| \lambda \|_1 \leq B.
\]
The parameter $B$ controls the fairness–performance trade-off: $B=0$ enforces the strictest fairness (often at the expense of accuracy), while larger values of $B$ progressively relax the constraints. In our experiments, we sweep $B$ over an exponential grid $\{0.01, 0.1, 1, 10, 100\}$ to generate a continuum of models analogous to the linear $\Theta$ sweep in FTC. In both datasets, we treat a single protected attribute (marital status for Bank Marketing, gender for Adult), so all fairness constraints apply to that variable. 

\paragraph{Metrics and Evaluation}\mbox{}\\
As described in Section~\ref{methodology}, each candidate model is characterized by user-specified predictive performance metrics (e.g.\ accuracy, ROC AUC, etc.), fairness metrics (e.g.\ strong demographic parity, equalized odds, etc.), and feature-importance measures (e.g.\ SHAP values, permutation importances, or alternative attribution methods).  
Cluster validity is assessed through the composite score in Equation~\ref{eq:composite}, and the optimal number of clusters is selected according to Equation~\ref{eq:kstar}.

\subsection{Results} \label{results}
We illustrate key results from our methodology using the Adult dataset with the Fair Tree Classifier (FTC) as a representative example. In our main analysis, we employ the \textbf{multi‐attribute} variant of FTC, which enforces simultaneous fairness across all protected dimensions by computing strong demographic parity (SDP) for each sensitive attribute (race, gender, age) and then taking the worst‐case SDP value as the model’s fairness score. 
Detailed results for additional setups are provided in Appendix~\ref{appendix_results}:  
FTC run in \textbf{single‐attribute} mode on Age, Gender, and Race in the Adult dataset, and FTC run in both \textbf{multi‐attribute} and \textbf{single‐attribute} (Age) modes in the Bank Marketing dataset.

\paragraph{Transformation Diagnostics}\mbox{}\\
After learning the Mahalanobis metric \(M = L^\top L\), we first inspect \(M\) itself to ensure it departs from the identity and exhibits meaningful off‐diagonal structure. Next, we verify that the transformation preserves local neighborhoods while reshaping global distances by plotting a “distance change” heatmap: 
\[
  \Delta d_{ij} = d_{ij}^{\mathrm{before}} - d_{ij}^{\mathrm{after}},
\]
with models ordered by increasing fairness penalty \(\theta\). Deep blue cells, which appear primarily off the main diagonal, show that ITML contracts originally distant pairs, whereas near‐white diagonal entries indicate minimal movement for already‐close pairs. This confirms that our embedding reshapes global relationships to reflect fairness–performance proximity while keeping local neighborhoods intact (see Appendix~\ref{heatmap}).

\begin{figure*}[ht]
    \centering
    \includegraphics[width=0.48\textwidth]{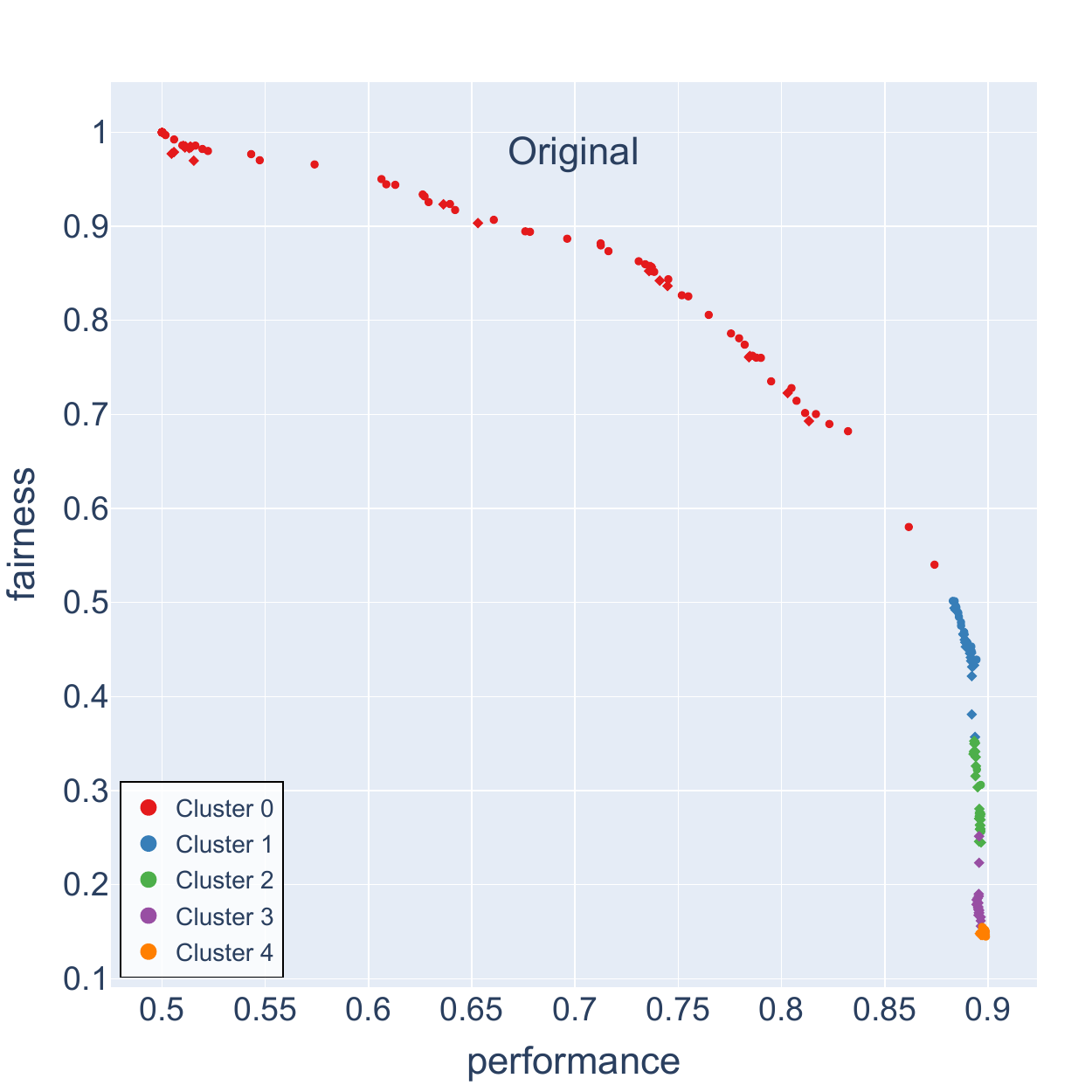}
    \hfill
    \includegraphics[width=0.48\textwidth]{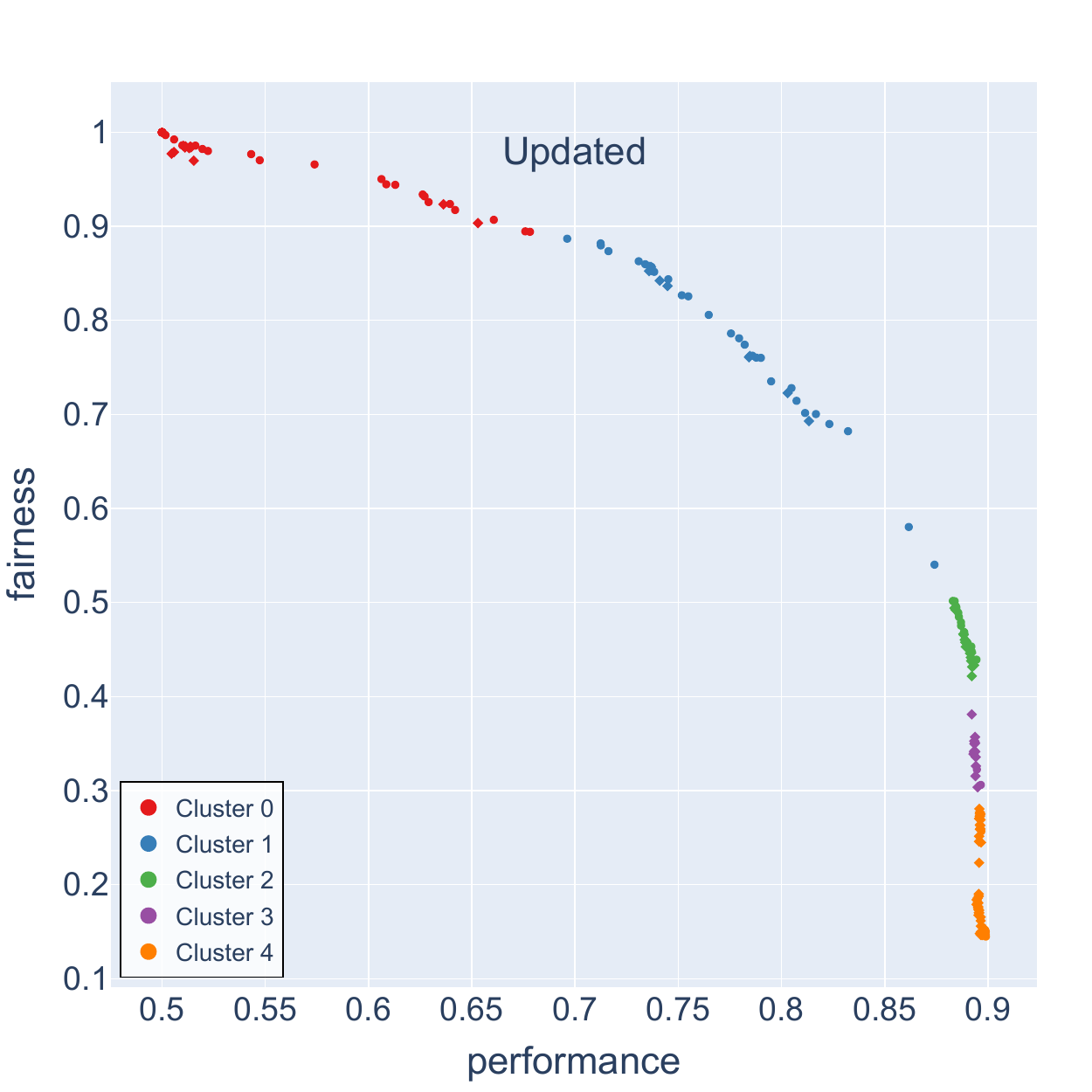}
    \caption{Side‐by‐side comparison of $k=5$ cluster assignments in fairness–performance space using (left) raw feature importances versus (right) Mahalanobis‐transformed importances. The learned metric reduces cluster overlap, making group boundaries much clearer.}
    \label{fig:pareto_fixed_side_by_side}
\end{figure*}

To visually isolate the specific impact of our learned Mahalanobis metric (independent of how $k$ was chosen), we now fix $k=5$ in both the raw and transformed feature-importance spaces. Figure~\ref{fig:pareto_fixed_side_by_side} shows side-by-side maps using the same $k$, directly comparing cluster overlap before and after metric learning.

\paragraph{Clustering and Model Grouping}\mbox{}\\
We determine the optimal number of clusters ($k$) using our composite validation approach. For the Adult dataset (FTC), composite validation scores identify an optimal $k = 5$ (Table~\ref{tab:composite_scores}).

\begin{table}[!ht]
    \centering
    \caption{Composite validation metrics for varying $k$ on FTC–Adult dataset (Top 5 by composite score)\label{tab:composite_scores}}
    \begin{tabular}{cccccc}
        \toprule
        \textbf{$k$} & 
        \textbf{Silhouette~($\uparrow$)} & 
        \textbf{Calinski–Harabasz~($\uparrow$)} & 
        \textbf{Davies–Bouldin~($\downarrow$)} & 
        \textbf{Dunn~($\uparrow$)} & 
        \textbf{Composite Score~($\uparrow$)} \\
        \midrule
         5  & 0.80750 &  4344.64  & 0.31823 & 0.17496 & \textbf{1.57026} \\ 
         3  & 0.81735 &  3001.11  & 0.24495 & 0.04214 & 0.70524      \\ 
         4  & 0.76866 &  5502.35  & 0.34604 & 0.02173 & 0.28692      \\ 
        19  & 0.59900 & 12639.69  & 0.44825 & 0.04806 & 0.23102      \\ 
        20  & 0.59943 & 13133.71  & 0.49047 & 0.05723 & 0.21167      \\ 
        \bottomrule
    \end{tabular}
\end{table}
\FloatBarrier
As shown in Table~\ref{tab:composite_scores}, our composite validation score peaks at $k=5$ with a value of \textbf{1.57026}, indicating that partitioning the FTC–Adult model space into five groups yields the best overall balance of cohesion and separation under our four criteria. Although the Silhouette (0.81735) and Davies–Bouldin (0.24495) indices both favour $k=3$, and the Calinski–Harabasz index reaches its maximum at $k=20$ (13 133.71), the Dunn index attains its highest value at $k=5$ (0.17496). By averaging the z-scored metrics, the composite score for $k=5$ substantially exceeds the next best configuration ($k=3$, 0.70524), validating a robust and stable choice of $k$ that balances compactness and separation across diverse criteria.

We further observe that each individual index exhibits known biases, as Calinski–Harabasz tends to increase monotonically with $k$ \cite{calinski_problem}, the Silhouette score suffers from shape bias \cite{Silhouettes}, Davies–Bouldin presumes equal cluster sizes and densities (reducing reliability on imbalanced or non‐spherical clusters) \cite{Davies}, and the Dunn index is highly sensitive to outliers \cite{dunn_problem}. By integrating all four measures into a single composite score, we leverage their complementary strengths while mitigating these individual drawbacks, resulting in a more reliable clustering choice.

\begin{figure*}[ht]
  \centering
  \includegraphics[width=0.7\textwidth]{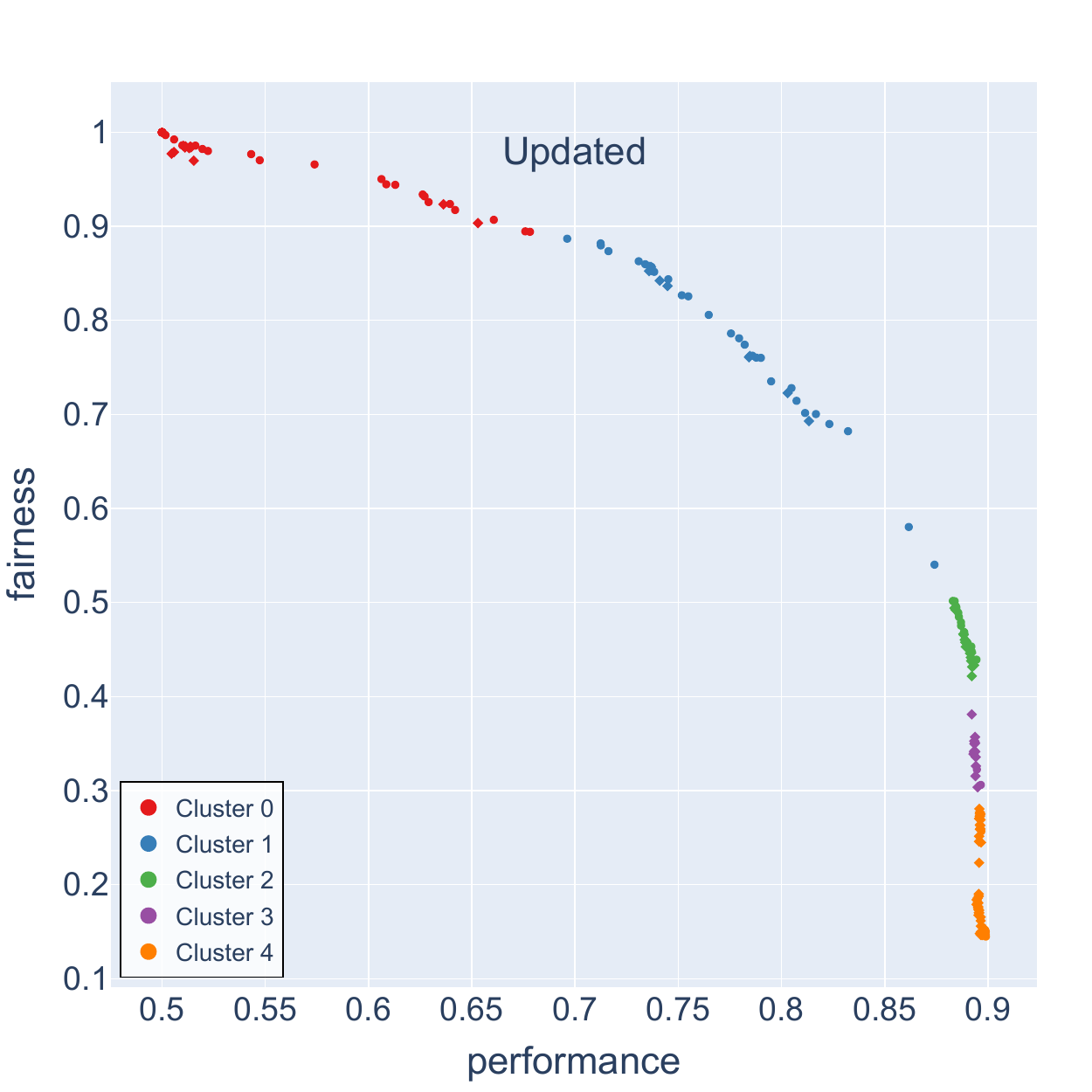}
  \caption{Cluster assignments in the Mahalanobis‐transformed feature‐importance space (optimal $k=5$), plotted in the fairness–performance plane. Distinct clusters highlight clear trade‐off archetypes.}
  \label{fig:pareto_transformed}
\end{figure*}

\paragraph{Clustered Fairness–Performance Map}\mbox{}\\
Using the optimal $k$ determined by our composite score, we visualize the final clustering in the Mahalanobis‐transformed feature‐importance space. Figure~\ref{fig:pareto_transformed} shows the models arranged in the fairness–performance plane, coloured by cluster membership. In this transformed space, models form clear, banded groups along the Pareto frontier: intra‐cluster distances are contracted for models sharing nearly identical fairness–performance balances, while inter‐cluster distances are expanded for those with divergent trade‐offs. This configuration yields well‐separated archetypes whose feature‐importance signatures directly align with stakeholder‐relevant criteria, greatly simplifying the selection of representative models.

It is worth mentioning that this clustering approach aims to create groups of models with similar fairness–performance trade-off characteristics. By clustering in the space of feature importances (rather than directly on fairness or accuracy metrics), we capture how different models achieve their results. For example, some models may achieve high performance by heavily weighting certain predictive features, while others may sacrifice using those features to satisfy fairness constraints. The ITML step ensured that the clustering is not overly sensitive to scale differences and that relevant variations in feature importance (which might correlate with fairness behavior) are taken into account.

\paragraph{Cluster Homogeneity and Trade-off Profiles}\mbox{}\\
Table~\ref{tab:cluster_metrics} summarizes, for each of the five clusters in the FTC–Adult model space:
\begin{itemize}
  \item \emph{Size} ($n_{\mathrm{points}}$): number of models in the cluster,
  \item \emph{Total variance}: sum of the individual feature importance variances computed as the sum of variances across all feature dimensions, i.e., $\sum_{j=1}^{P}\mathrm{Var}(X_j)$, where $X_j$ represents the importance values of the $j^{th}$ feature for models within the cluster. This metric indicates the overall spread and homogeneity of feature attribution patterns within each cluster,
  \item \emph{Mean fairness} and \emph{mean performance}: average fairness and performance of the models in the respective cluster(±SD).
\end{itemize}
Lower total variance implies more homogeneous feature‐importance profiles within a cluster, while the fairness/performance averages locate each group along the accuracy–fairness frontier.

\begin{table}[h!]
\centering
\caption{Cluster metrics with total variance, Performance (mean ROC AUC±SD), and Fairness (mean SDP±SD).\label{tab:cluster_metrics}}
\begin{tabular}{ccccc}
\toprule
\textbf{Cluster} & \textbf{$n_{\mathrm{points}}$} & \textbf{Total Variance} & \textbf{Performance (±SD)} & \textbf{Fairness (±SD)} \\
\midrule
0 & 59 & 0.000016 & $0.5489 \pm 0.0820$ & $0.9621 \pm 0.0815$ \\
1 & 36 & 0.000027 & $0.7649 \pm 0.0356$ & $0.7979 \pm 0.0645$ \\
2 & 36 & 0.000086 & $0.8891 \pm 0.0030$ & $0.4614 \pm 0.0214$ \\
3 & 15 & 0.000308 & $0.8938 \pm 0.0010$ & $0.3376 \pm 0.0198$ \\
4 & 54 & 0.000112 & $0.8964 \pm 0.0011$ & $0.1953 \pm 0.0500$ \\
\bottomrule
\end{tabular}
\end{table}

Table~\ref{tab:cluster_metrics} reveals a pronounced stratification of model behaviours along the fairness–performance frontier. The first two clusters exhibit very low internal variance and represent stable strategies: \emph{Cluster 0} achieves near‐perfect fairness with minimal predictive power, while \emph{Cluster 1} balances high fairness with moderate accuracy. \emph{Clusters 2 through 4} trace a smooth progression toward greater accuracy at the expense of fairness, with \emph{Cluster 3} standing out for its notably higher feature‐importance heterogeneity, indicating that comparable trade‐offs can be reached via diverse feature‐use patterns, whereas \emph{Cluster 4} consistently attains the highest accuracy and lowest fairness with relatively uniform profiles. Together, these summaries offer stakeholders both clear trade‐off points and insight into the consistency of underlying model behaviors.  

\begin{figure*}[!htbp]
  \centering

  \begin{subfigure}[t]{0.48\textwidth}
    \includegraphics[width=\textwidth]{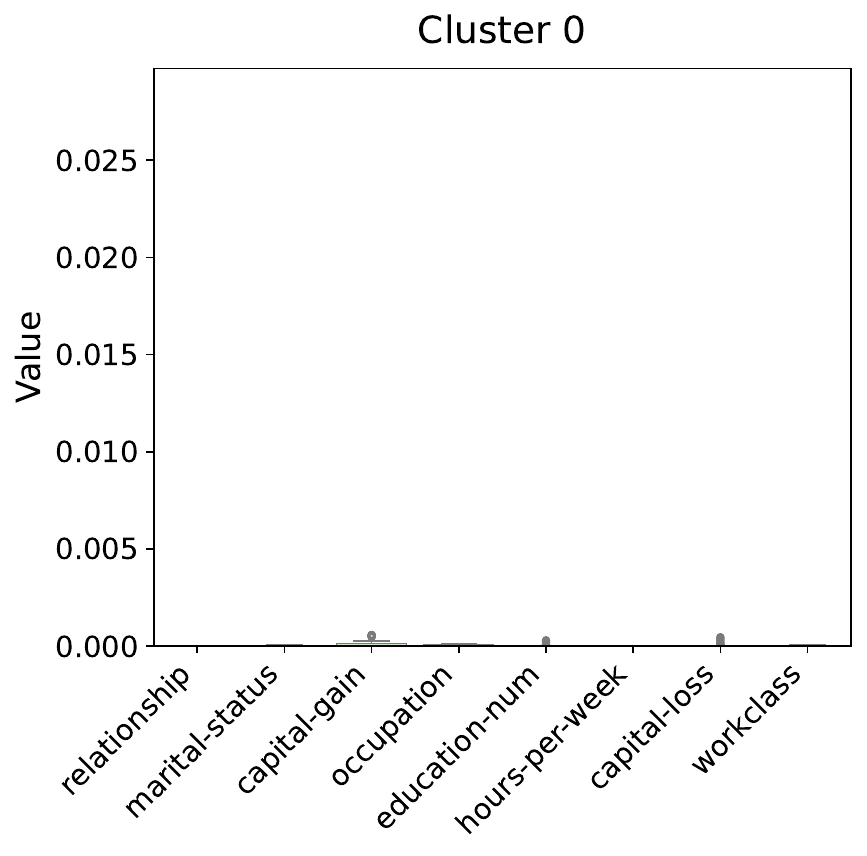}
    \label{fig:box_cluster0}
  \end{subfigure}
  \hfill
  \begin{subfigure}[t]{0.48\textwidth}
    \includegraphics[width=\textwidth]{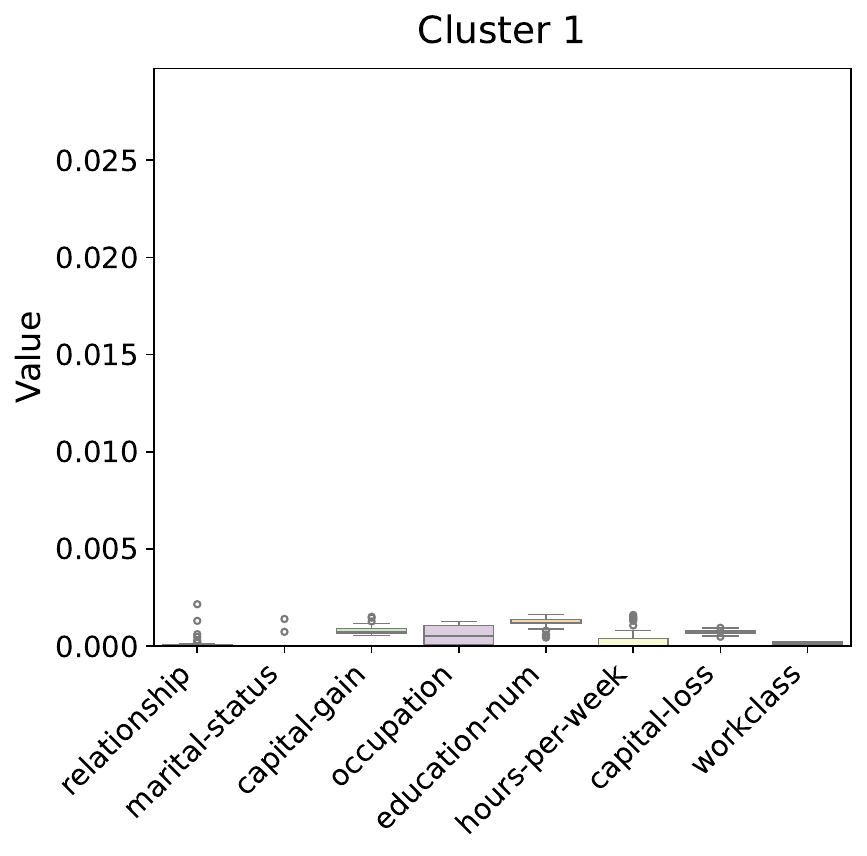}
    \label{fig:box_cluster1}
  \end{subfigure}

  \vspace{1ex}

  \begin{subfigure}[t]{0.48\textwidth}
    \includegraphics[width=\textwidth]{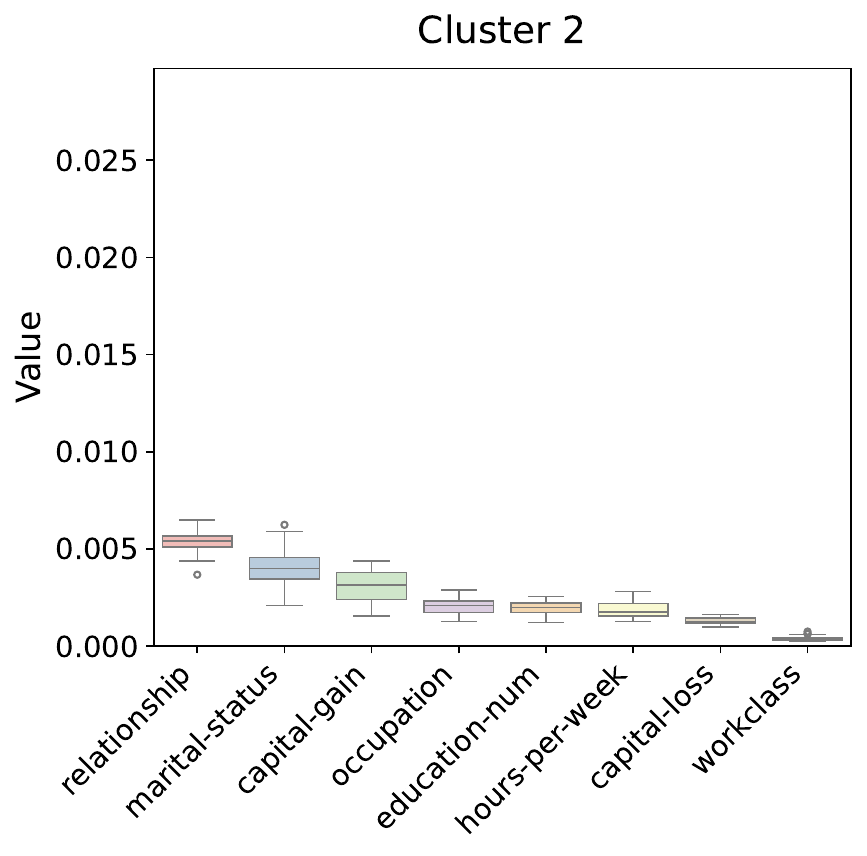}
    \label{fig:box_cluster2}
  \end{subfigure}
  \hfill
  \begin{subfigure}[t]{0.48\textwidth}
    \includegraphics[width=\textwidth]{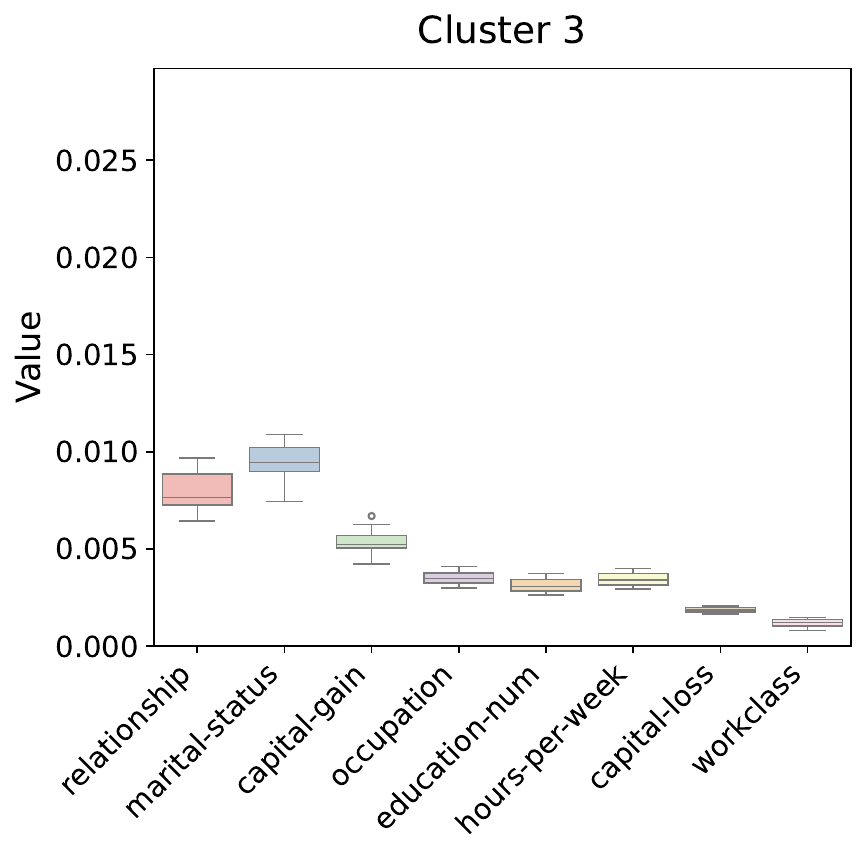}
    \label{fig:box_cluster3}
  \end{subfigure}

  \vspace{1ex}

  \begin{subfigure}[t]{0.48\textwidth}
    \includegraphics[width=\textwidth]{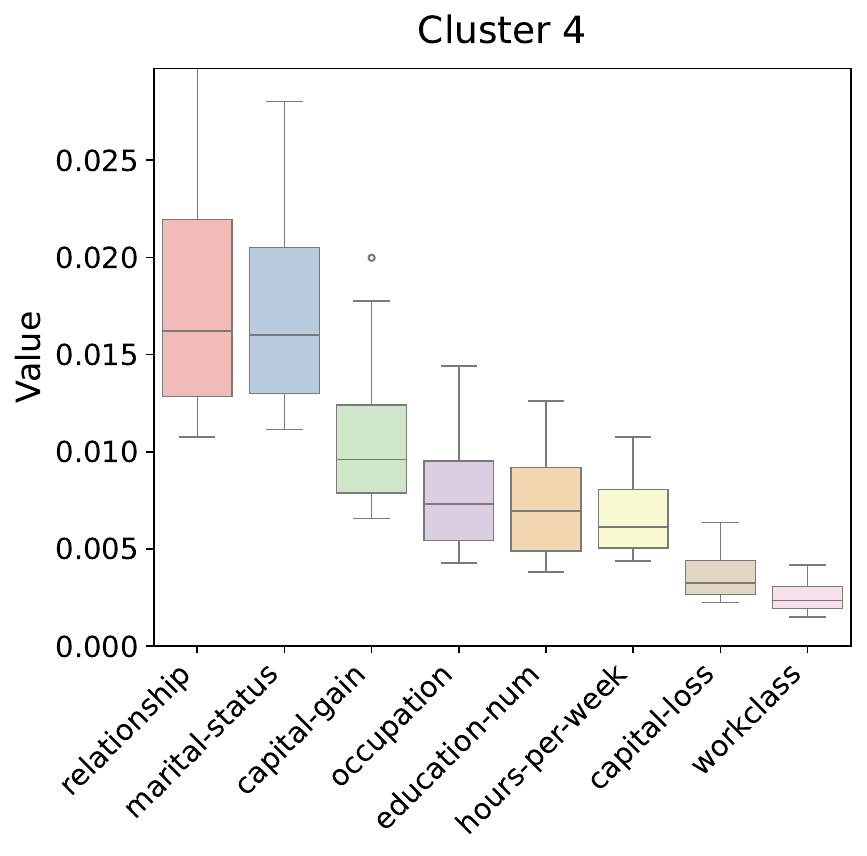}
    \label{fig:box_cluster4}
  \end{subfigure}

  \caption{%
    Per‐cluster boxplots of the top eight average SHAP feature importances.
    Each subfigure shows one cluster, with identical vertical scales
    to facilitate cross‐cluster comparison.%
  }
  \label{fig:all_clusters_boxplots}
\end{figure*}

\paragraph{Cluster-Level Feature Attribution}\mbox{}\\
Figure~\ref{fig:all_clusters_boxplots} presents side‐by‐side boxplots of the same set of nine features—ordered by their overall mean SHAP importance across all clusters (relationship, marital‐status, capital‐gain, occupation, education‐num, hours‐per‐week, capital‐loss, workclass), so that relative differences in feature use are directly comparable.

The boxplots reveal a smooth progression in feature reliance along the fairness–accuracy continuum. In cluster 0, models under the strictest fairness constraints exhibit near-zero SHAP values across all nine predictors, effectively reducing their behavior to a nearly constant-output strategy that avoids any meaningful feature influence. This phenomenon aligns with theoretical results showing that perfect fairness criteria can force classifiers into trivial, constant predictions to eliminate disparity \cite{pinzón2021impossibilitynontrivialaccuracyfairness}. In other words, these models guarantee perfect fairness by making nearly identical predictions for everyone, but in doing so they lose almost all ability to distinguish between different outcomes.

The next cluster shows modest yet consistent emphasis on educational and economic indicators, notably education-num and capital-gain, reflecting cautious use of those features under moderate fairness constraints. As models begin to trade off more accuracy for fairness, socio-demographic attributes such as relationship and marital-status assume greater median importance and display wider dispersion, with occupation and education-num contributions also broadening, a sign of more varied feature-use patterns emerging in these mid-range accuracy clusters. At the accuracy extreme, models amplify the influence of relationship, marital-status, and capital-gain, and uniquely integrate gender and age, two sensitive attributes, into their predictive processes. This reliance on these attributes, while boosting predictive performance, raises critical questions about disparate treatment and underscores the necessity analysis. Together, these evolving feature-importance signatures illuminate how different fairness–performance trade-offs leave distinct imprints on model behavior, guiding stakeholders toward archetypal classifiers that best align with their ethical and operational priorities.  

\FloatBarrier

\subsection{Fairness–Performance Trade‐off Implications for Stakeholders}

Below we characterize five model archetypes, each corresponding to a distinct segment of the fairness–performance spectrum and differing in the coherence of their feature‐importance profiles (see Table~\ref{tab:cluster_metrics}, Figure~\ref{fig:pareto_transformed}, and Figure~\ref{fig:all_clusters_boxplots}). By focusing on these archetypal groups, stakeholders such as regulatory bodies, ethics review boards, or model‐selection committees can identify model sets that align with their specific fairness–performance requirements, without needing to evaluate each individual model.

\paragraph{Maximal Fairness, Minimal Predictive Power Archetype (Cluster 0).}\mbox{}\\
Cluster 0 achieves maximal SDP ($\approx0.9621\pm0.0815$) by uniformly ignoring input features, yielding trivial classifiers (ROC AUC $\approx0.5489\pm0.0820$) that exhibit near-zero SHAP values across all predictors. Such models satisfy stringent fairness criteria but provide almost no discriminatory capability.

\noindent
\emph{Stakeholder takeaway}: Use these models only under strict regulatory or moral imperatives that prioritize parity over predictive nuance, recognizing their limited practical utility. Practically, they might be akin to always predicting the majority class or random guessing with a fixed probability for positive outcome.

\paragraph{Fairness-Centric with Moderate Predictive Utility Archetype (Cluster 1).}\mbox{}\\
Cluster 1 occupies a critical intermediate position, achieving moderate accuracy (mean of 0.7649 ± 0.0356) combined with significantly higher fairness (mean of 0.7979 ± 0.0645). With notably low internal variance (0.000027), the models within this cluster show consistent, stable behavior. They emphasize economic indicators, especially \textit{capital-gain}, \textit{education-num}, \textit{capital-loss}, and \textit{hours-per-week}, presenting reliable predictors while consistently maintaining a strong fairness profile. 

\noindent
\emph{Stakeholder takeaway}: This cluster provides an ideal choice for stakeholders who require substantial fairness without excessively compromising predictive accuracy. Models in this archetype suit scenarios like equitable hiring processes, credit assessments, or other applications demanding both transparency and reasonable predictive performance.

\paragraph{Balanced Accuracy–Fairness Archetype (Clusters 2 and 3).}\mbox{}\\
Clusters 2 and 3 maintain near-peak ROC AUCs ($\approx0.8891\pm0.0030$ and $\approx0.8938\pm0.0010$) while improving SDP to moderate levels ($\approx0.4614\pm0.0214$ and $\approx0.3376\pm0.0198$). Their higher variances (0.000086 and 0.000308) reflect more diverse feature-use patterns: socio-demographic attributes (\textit{marital-status, relationship}) dominate, while \textit{occupation} and \textit{education-num} exhibit broader distributions. These archetypes strike a balanced compromise between fairness gains and strong predictive power.

\noindent
\emph{Stakeholder takeaway}: Ideal when slight accuracy reductions are tolerable in exchange for meaningful fairness improvements, though socio-demographic biases should be monitored further.

\paragraph{Max‐Accuracy Archetype (Cluster 4).}\mbox{}\\
Cluster 4 delivers the highest predictive performance ($\approx0.8964\pm0.0011$) at the expense of the lowest SDP ($\approx0.1953\pm0.0500$). Models within this group exhibit moderate internal heterogeneity, reflected by a total feature-importance variance of 0.000112. They predominantly utilize features like \textit{relationship}, \textit{marital-status}, \textit{capital-gain}, and \textit{education-num}.

\noindent
\emph{Stakeholder takeaway}: Stakeholders considering this cluster should recognize the significant accuracy benefits but must remain cautious regarding fairness implications, employing additional measures to manage and mitigate potential biases.

\bigskip
In summary, our clustering framework transforms a large, hard-to-manage set of fairness-aware models into five actionable archetypes. Decision makers can now directly map their requirements, whether maximal accuracy, moderate fairness, or strict parity, onto one of these clusters, dramatically streamlining the model‐selection process in high‐stakes settings.

\section{Conclusions and Outlook}
In this paper, we have presented an end-to-end framework for visual model selection that combines weakly supervised metric learning with feature‐importance clustering to illuminate the fairness–performance landscape of candidate classifiers. By learning a Mahalanobis embedding aligned with stakeholder‐relevant trade‐offs and applying a composite validation strategy to identify the optimal number of clusters, our method distills large Rashomon sets into a small number of archetypal model groups. Each archetype is characterized by its average accuracy and fairness scores, as well as a distinctive feature‐importance signature, enabling decision makers to rapidly pinpoint models that best match their operational and ethical priorities. Empirical results on the Adult and Bank Marketing datasets demonstrate that our approach both clarifies how fairness constraints reshape feature reliance and substantially reduces the cognitive burden of model selection.

However, our composite score approach for determining the optimal number of clusters also has inherent limitations that require further study. Although integrating multiple cluster validation indices reduces reliance on a single metric, the selection of indices and their equal weighting could introduce unintended biases. Further research is needed to systematically explore alternative weighting schemes, evaluate additional or different clustering metrics, and assess the sensitivity of the composite approach across a broader array of datasets and problem contexts.

Looking forward, our framework stands to benefit from real‐world deployment and structured user studies with domain experts, auditors, and policymakers on operational datasets, such as credit scoring, hiring, or criminal risk assessment, to validate its usability and effectiveness beyond benchmark settings. Simultaneously, expanding the library of fairness‐aware learners to encompass adversarial debiasing approaches, counterfactual fairness models, and post‐processing strategies like equalized‐odds adjustments will shed light on how diverse mitigation techniques manifest in the clustered embedding and feature‐importance signatures. By pursuing these directions, we aim to bridge the gap between fairness research and real‐world decision support, empowering stakeholders to make principled, transparent choices when deploying algorithmic systems in high‐stakes contexts.  

\begin{acknowledgments}
  This research is funded by the ICAI lab AI4Oversight.
  \url{https://www.ai4oversight.nl}.  
\end{acknowledgments}

\section*{Declaration on Generative AI}
During the preparation of this work, the authors used ChatGPT (OpenAI) to:
\emph{Improve writing style} and \emph{Paraphrase and reword sentences} (including minor grammar and spelling suggestions), in line with the CEUR-WS GenAI Usage Taxonomy.

\bibliography{main}

\appendix
\newpage

\section{Additional Experimental Results}\label{appendix_results}

In this appendix, we provide comprehensive additional results supporting the analyses presented in the main text. Specifically, we illustrate further details and visualizations for the Fair Tree Classifier (FTC) and Fair Logistic Regression (FLR), demonstrating their behavior on the Adult and Bank Marketing datasets under various fairness attribute considerations.

\subsection{Fair Tree Classifier (FTC)}
\subsubsection{Adult Dataset: Distance Change Heatmap}
\label{heatmap}

\begin{figure}[!htb]
  \captionsetup{type=figure}  
  \centering
  \includegraphics[width=1\linewidth]{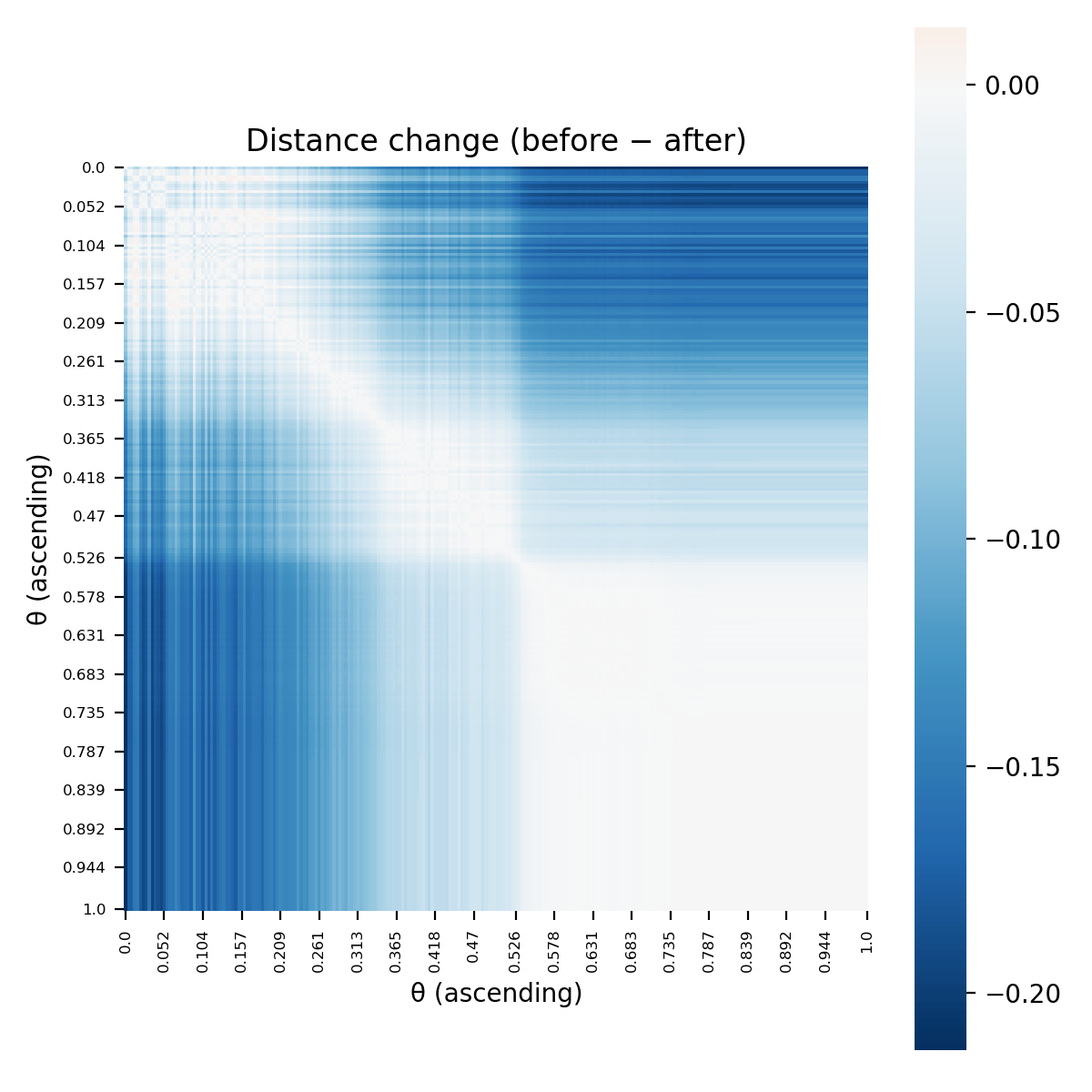}
  \caption{Heatmap of pairwise distance change \(\Delta d_{ij} = d_{ij}^{\mathrm{before}} - d_{ij}^{\mathrm{after}}\) for FTC on the Adult dataset, sorted by increasing \(\theta\). Deep blues show where ITML strongly contracts distances among similar models.}
  \label{fig:distance_change_adult}
\end{figure}

Figure \ref{fig:distance_change_adult} shows the pairwise distance change heatmap, highlighting how metric learning via ITML contracts distances among models with similar fairness–performance trade-offs.

\subsubsection{Single-attribute FTC (Adult Dataset)}
\begin{figure*}[!ht]
  \centering
  \begin{subfigure}[t]{0.48\textwidth}
    \includegraphics[width=\linewidth]{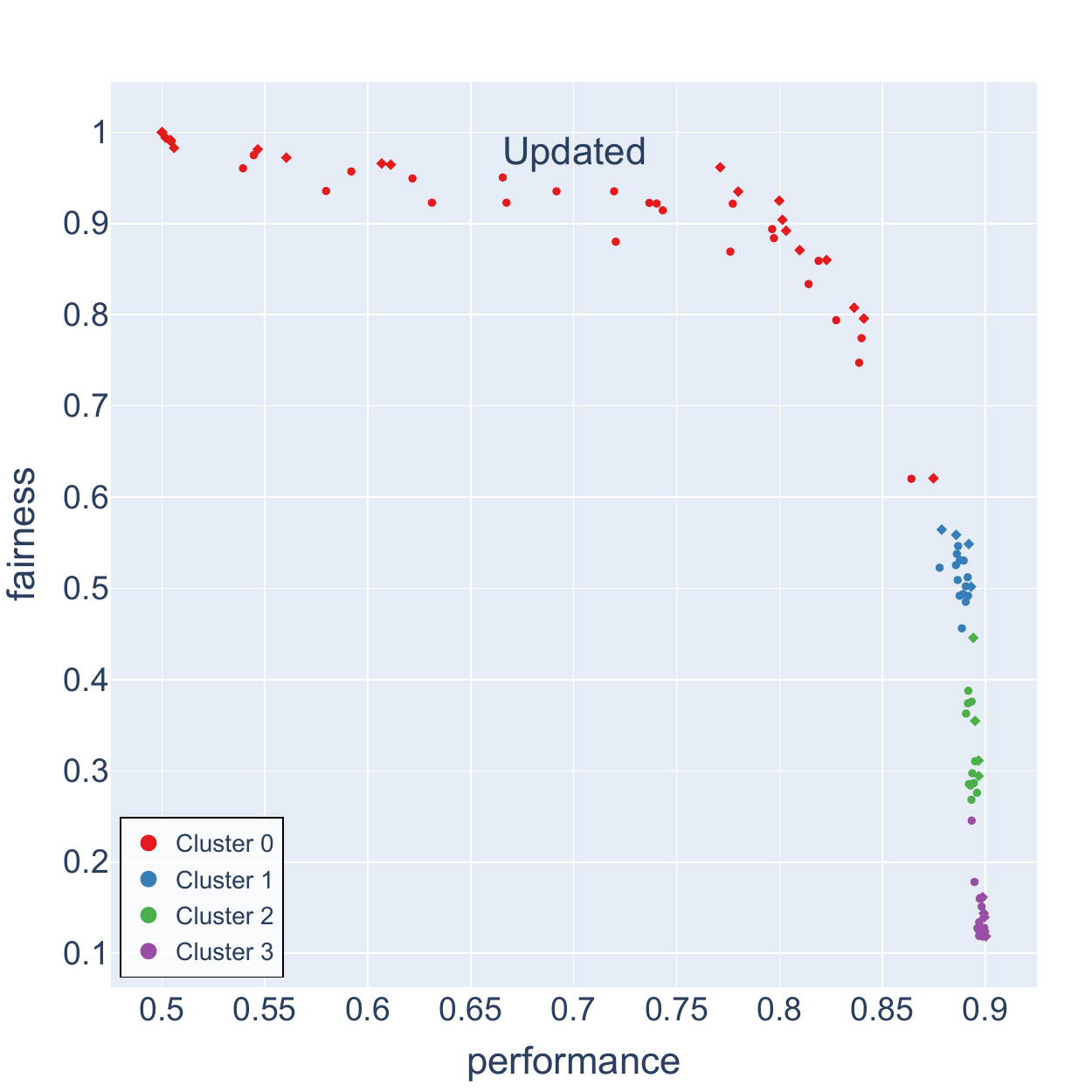}
    \caption{Age only}
    \label{fig:pareto_age_ftc}
  \end{subfigure}
  \hfill
  \begin{subfigure}[t]{0.48\textwidth}
    \includegraphics[width=\linewidth]{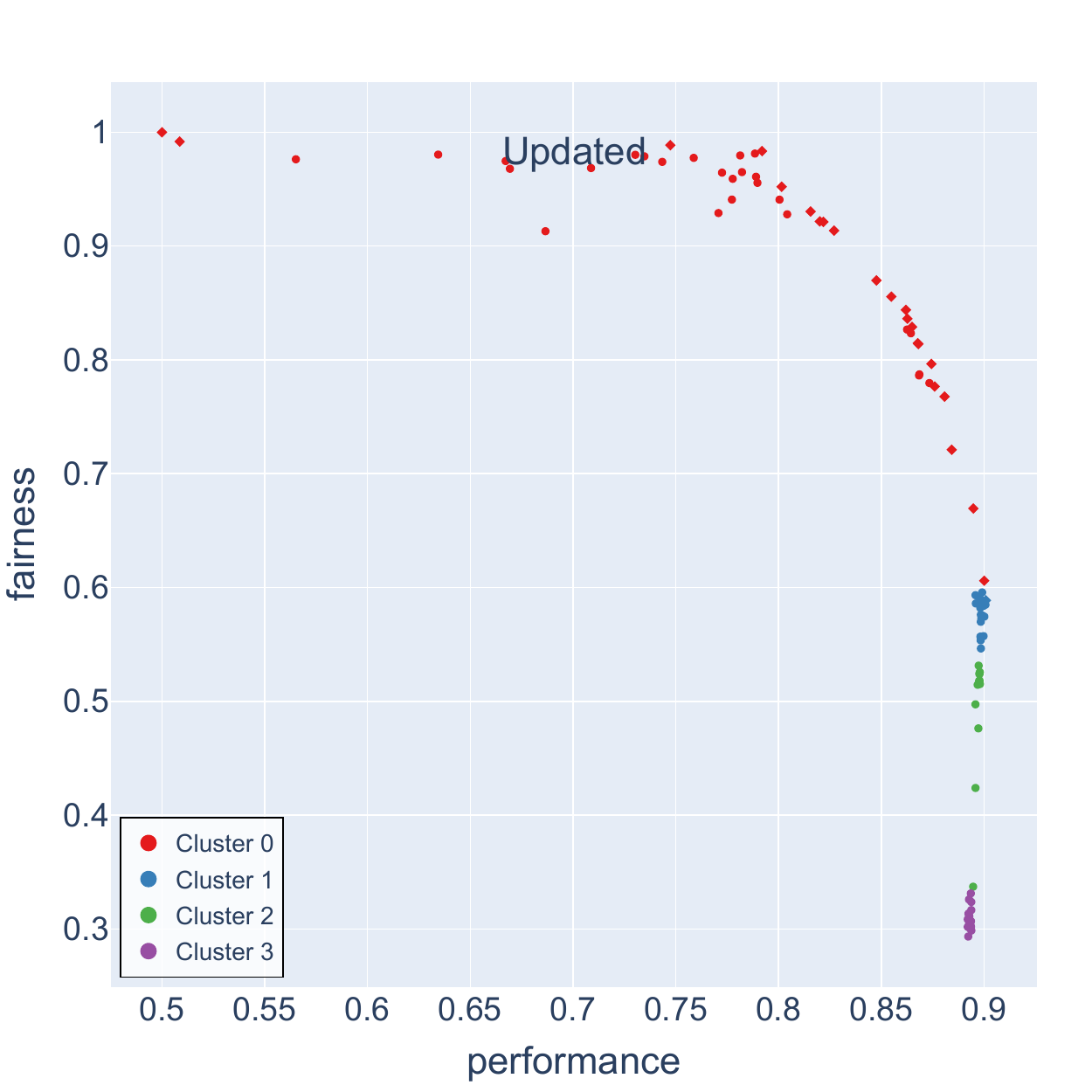}
    \caption{Gender only}
    \label{fig:pareto_gender_ftc}
  \end{subfigure}

  \vspace{1em}

  \begin{subfigure}[t]{0.5\textwidth}
    \centering
    \includegraphics[width=\linewidth]{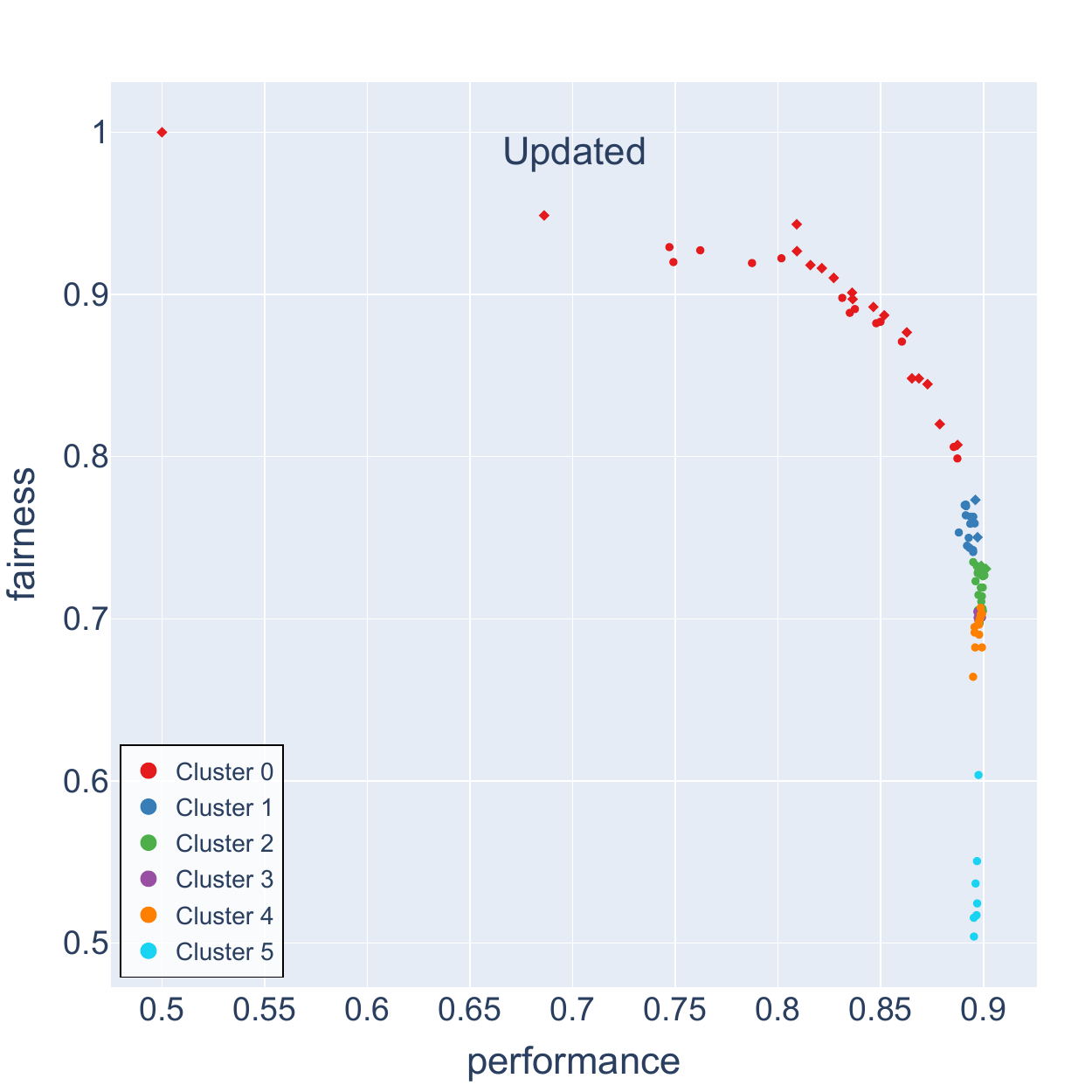}
    \caption{Race only}
    \label{fig:pareto_race_ftc}
  \end{subfigure}

  \caption{Single‐attribute FTC on Adult dataset: clustered model distributions when only (a) Age, (b) Gender, or (c) Race is treated as the protected attribute. Each uses its own optimal \(k\) from composite validation.}
  \label{fig:pareto_single_attr_adult_ftc}
\end{figure*}

To illustrate how enforcing fairness on a single sensitive attribute reshapes the model clusters, we re-ran FTC separately on \emph{Age}, \emph{Gender}, and \emph{Race}, each time fixing \(k\) via the composite score. Figure~\ref{fig:pareto_single_attr_adult_ftc} shows the resulting maps in fairness-performance space for the three runs.

\FloatBarrier

\newpage

\begin{table}[!t]
  \centering
  \caption{Composite validation metrics for single‐attribute FTC on Adult (top‐5 by composite).}
  \label{tab:comp_single_attr_adult_ftc}
  \scriptsize
  \begin{subtable}[t]{0.32\linewidth}
    \centering
    \caption{Gender only}
    \begin{tabular}{rcccc}
      \toprule
      $k$ & Sil~($\uparrow$) & CH~($\uparrow$) & DB~($\downarrow$) & Comp~($\uparrow$) \\
      \midrule
       4 & 0.7114 & 1037.4 & 0.4147 & \textbf{1.3308}\\
       18 & 0.5596 &   3390.4 & 0.4570 & 0.7769\\
      3 & 0.7255 & 747.4 & 0.3748 & 0.7725\\
      16 & 0.5748 & 2999.3 & 0.4618 & 0.43256\\
      17 & 0.5589 & 3064.95 & 0.4628 & 0.2624\\
      \bottomrule
    \end{tabular}
  \end{subtable}
  \hfill
  \begin{subtable}[t]{0.32\linewidth}
    \centering

    \caption{Age only}
    \begin{tabular}{rcccc}
      \toprule
      $k$ & Sil~($\uparrow$) & CH~($\uparrow$) & DB~($\downarrow$) & Comp~($\uparrow$) \\
      \midrule
       4 & 0.7191 &  943.47 & 0.4223& \textbf{1.0312}\\
       20 & 0.6265 &  4419.40 & 0.4170 & 0.8724\\
       7 & 0.6191 & 1013.0 & 0.6015 & 0.7482\\
       18 & 0.6607 &  899.2 & 0.5317 & 0.6394\\
      14 & 0.4887 & 1119.4 & 0.7272 & 0.0374\\
      \bottomrule
    \end{tabular}
  \end{subtable}
  \hfill
  \begin{subtable}[t]{0.32\linewidth}
    \centering
    \caption{Race only}
    \begin{tabular}{rcccc}
      \toprule
      $k$ & Sil~($\uparrow$) & CH~($\uparrow$) & DB~($\downarrow$) & Comp~($\uparrow$) \\
      \midrule
       6 & 0.5776 &  882.8 & 0.5808 & \textbf{0.6134}\\
       7 & 0.6226 &  839.0 & 0.5405 & 0.5509\\
       5 & 0.6477 & 755.8 & 0.4509 & 0.4969\\
       8 & 0.6174 &  790.0 & 0.5144 &  0.4877\\
      18 & 0.4489 & 933.1 & 0.6894 &  0.0448\\
      \bottomrule
    \end{tabular}
  \end{subtable}
\end{table}

\noindent\textbf{Key insights.}
\begin{itemize}
  \item \textbf{Age‐only fairness} yields \(k^*=4\), producing four well‐spaced archetypes that smoothly trade off fairness and performance, similar to the multi‐attribute case but with slightly tighter groupings, reflecting the more limited disparity introduced by age.
  \item \textbf{Gender‐only fairness} also selects \(k^*=4\), but the fairness span is wider, from near‐perfect parity down to substantially reduced fairness, indicating sharper trade‐offs when constraining on gender alone.
  \item \textbf{Race‐only fairness} requires \(k^*=6\), revealing a more intricate landscape: six distinct clusters capture nuanced shifts in accuracy‐parity balances across racial groups, suggesting stakeholders need finer granularity when race is the protected attribute.
\end{itemize}

\newpage
\subsubsection{Bank‐Marketing Dataset}

To demonstrate FTC’s behavior on the Bank‐Marketing dataset, we show both the multi‐attribute run (worst‐case SDP) and the age‐only run side by side. Each subfigure uses its own optimal \(k\) from composite validation.

\begin{figure}[!htb]
  \centering
  \begin{subfigure}[t]{0.48\textwidth}
    \includegraphics[width=\linewidth]{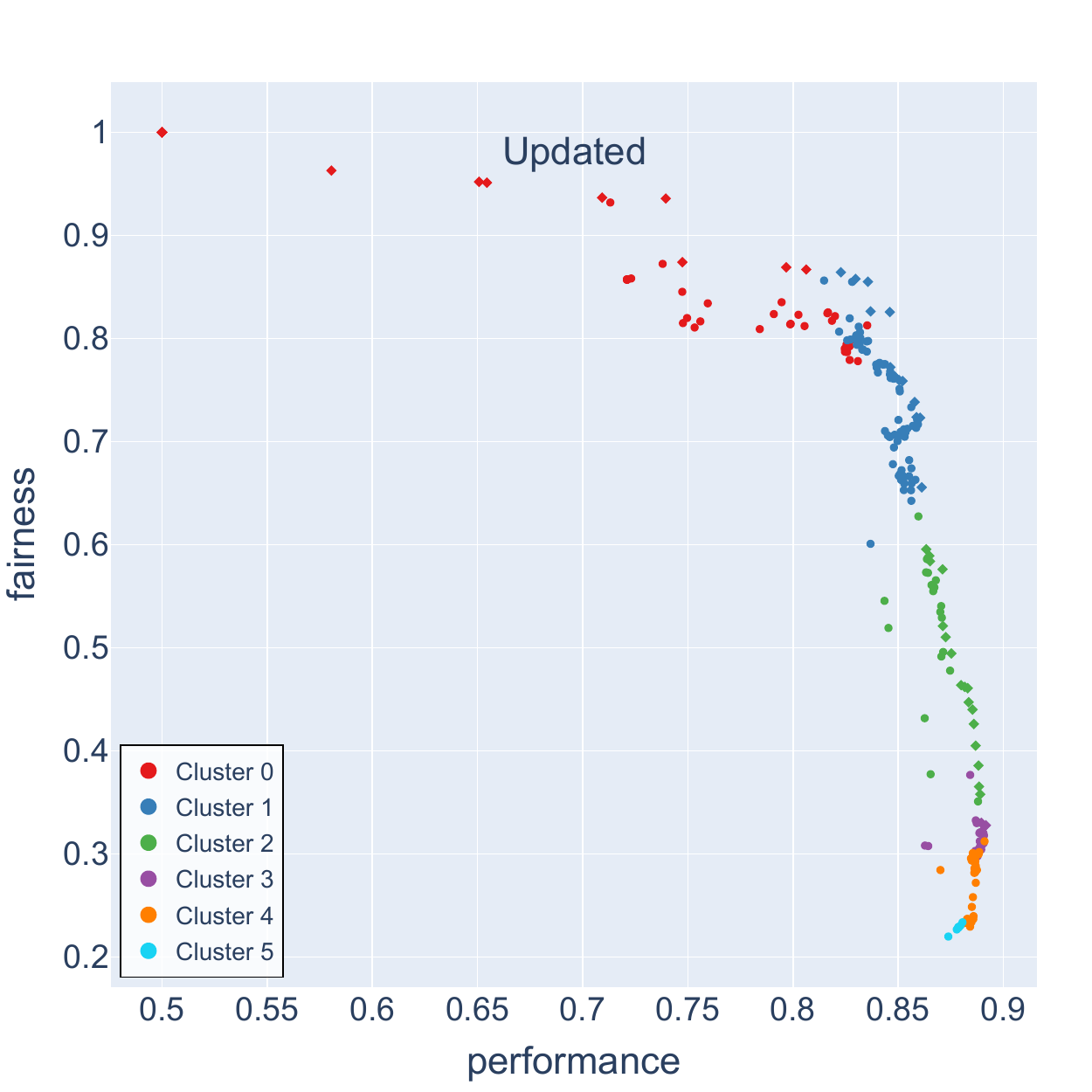}
    \caption{Multi‐attribute FTC (optimal \(k=6\)).}
    \label{fig:pareto_bank_multi_ftc}
  \end{subfigure}
  \hfill
  \begin{subfigure}[t]{0.48\textwidth}
    \includegraphics[width=\linewidth]{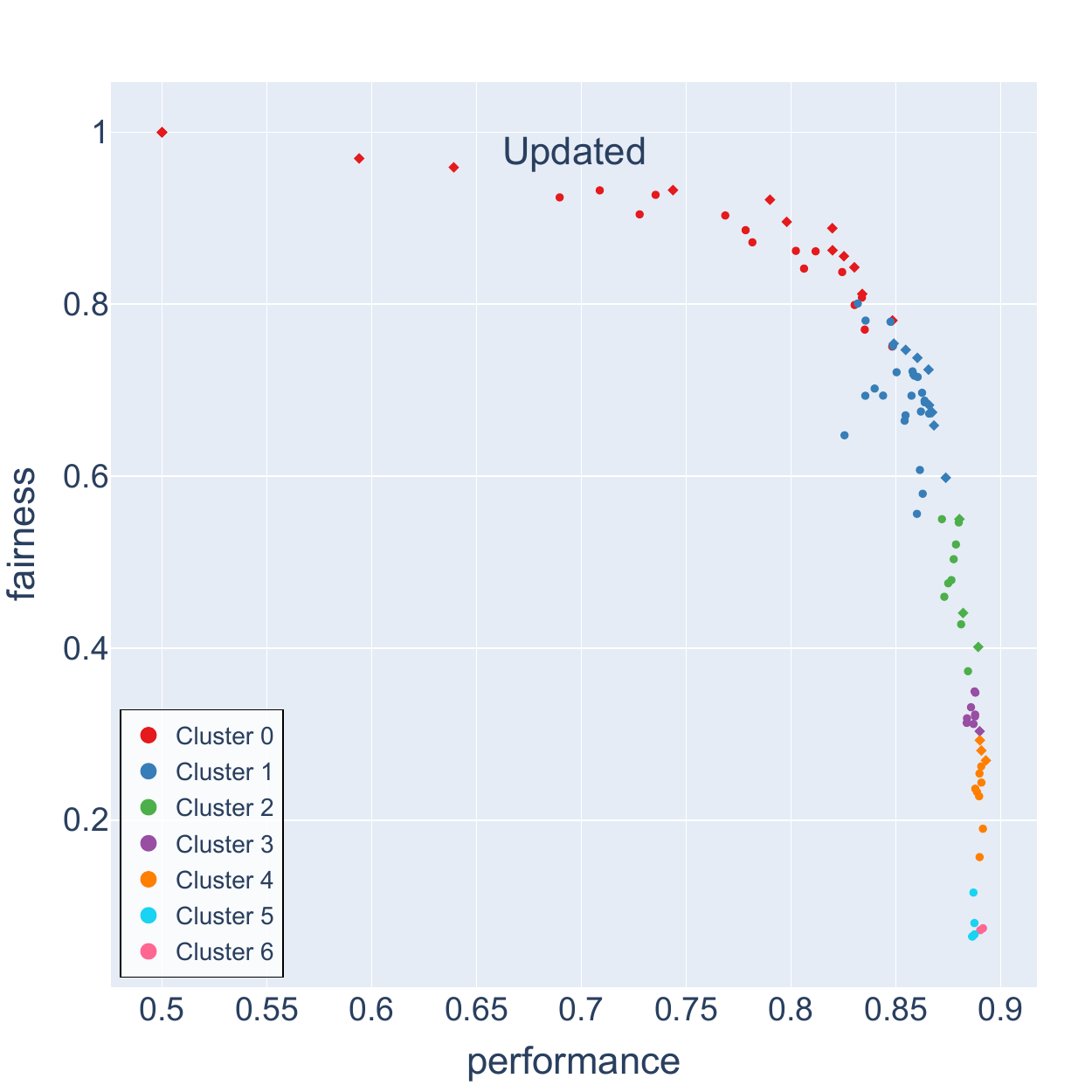}
    \caption{Age‐only FTC (optimal \(k=7\)).}
    \label{fig:pareto_bank_age_ftc}
  \end{subfigure}
  \caption{FTC clustering on the Bank‐Marketing dataset under (a) multi‐attribute fairness and (b) age‐only fairness.}
  \label{fig:pareto_bank_ftc_combined}
\end{figure}
\begin{table}[!ht]
  \centering
  \caption{FTC–Bank Marketing composite scores (top‐5 by composite).}
  \label{tab:comp_bank_ftc_combined}
  \scriptsize
  \begin{tabular}{r|cccc|r|cccc}
    \toprule
    & \multicolumn{4}{c|}{\textbf{Multi‐attribute}} 
    & & \multicolumn{4}{c}{\textbf{Age only}} \\
    \(k\) & Sil ↑ & CH ↑ & DB ↓ & Comp ↑ 
      & \(k\) & Sil ↑ & CH ↑ & DB ↓ & Comp ↑ \\
    \midrule
     6 & 0.6655 & 2242.9 & 0.4091 & \textbf{0.9113} 
     & 7 & 0.6649 &  1401.93 & 0.3981 & \textbf{0.6112} \\
    16 & 0.6103 & 4881.6 & 0.4830 & 0.6663      
     & 18 & 0.5233 &  1834.11 & 0.4084 & 0.4988      \\
     8 & 0.6555 & 2468.9 & 0.4471 & 0.4826      
     & 15 & 0.5270 &  1339.24 & 0.4121 & 0.4861      \\
    13 & 0.6257 & 4036.2 & 0.4571 & 0.4575      
     & 16 & 0.5240 &  1492.73 & 0.4678 & 0.4752      \\
    12 & 0.6348 & 3838.7 & 0.4583 & 0.3512      
     & 20 & 0.5150 &  2029.74 & 0.4410 & 0.4207      \\
    \bottomrule
  \end{tabular}
\end{table}

\noindent\textbf{Key insights.}
\begin{itemize}
  \item \textbf{Different cluster granularity:} Under multi‐attribute fairness, FTC yields \(k^*=6\) archetypes, whereas enforcing only age produces \(k^*=7\). Enforcing a single attribute allows finer distinctions along the fairness–performance continuum, revealing subtler trade‐off patterns that merge when multiple attributes are considered jointly.
  \item \textbf{Cluster overlap in age‐only run:} In Figure~\ref{fig:pareto_bank_age_ftc}, clusters 0 and 1 partially overlap at the high‐fairness/high‐performance end, and analogous overlaps occur between clusters 3–4 and 4–5. These overlaps suggest that some models exhibit very similar fairness–performance profiles despite belonging to different clusters. This is an inherent artifact of spherical k-means in a complex, high‐dimensional feature‐importance space \cite{spherical_clustering}, and it highlights potential benefits of exploring alternative clustering methods (e.g.\ Gaussian mixture models, density‐based clustering) or refining metric‐learning constraints to further sharpen cluster separations.
  \item \textbf{Smooth trade‐off progression:} Despite overlaps, both settings reveal a coherent ordering of clusters along the Pareto frontier, from maximal fairness (low accuracy) to maximal accuracy (low fairness). Stakeholders can still identify representative archetypal groups, recognizing that some clusters may share boundary models with highly similar behaviors.
\end{itemize}

\subsection{Fair Logistic Regression (FLR)}

\subsubsection{Adult Dataset}
To evaluate how FLR navigates the fairness–performance trade-off on \emph{Adult} by using gender as sensitive attribute, we clustered models using SHAP feature importances under the optimal \(k\) determined by composite validation.  

\begin{figure}[!htb]
  \centering
\includegraphics[width=0.6\linewidth]{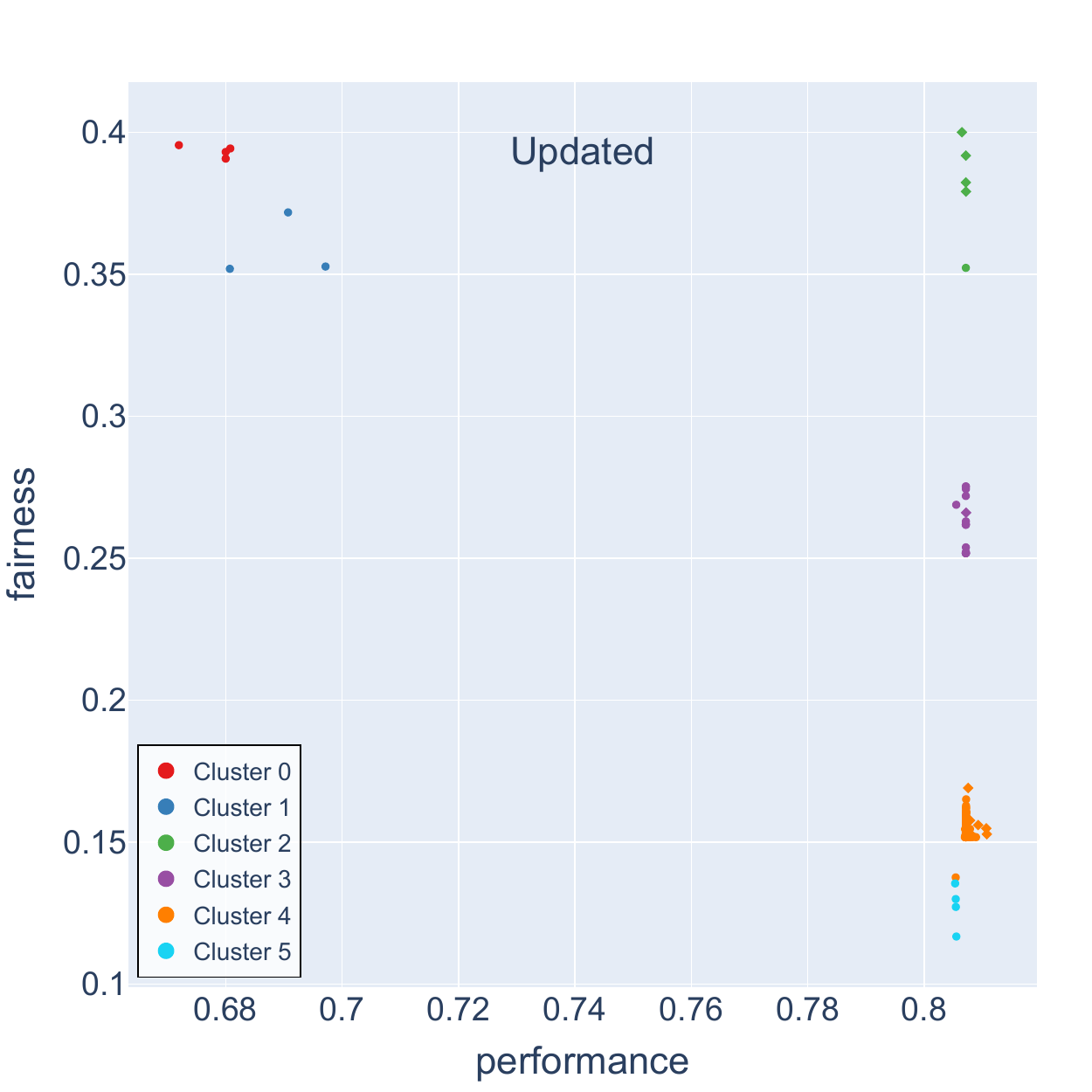}
  \caption{Fair Logistic Regression on the Adult dataset: clustered model distributions in fairness-performance space for optimal \(k=6\).}
  \label{fig:pareto_flr_adult}
\end{figure}

\begin{table}[!ht]
  \centering
  \scriptsize
  \caption{Composite validation metrics for FLR–Adult (top-5 by composite).}
  \label{tab:comp_flr_adult}
  \begin{tabular}{rcccc}
    \toprule
    \(k\) & Sil~($\uparrow$) & CH~($\uparrow$) & DB~($\downarrow$) & Comp~($\uparrow$) \\
    \midrule
     6 & 0.9443 & 1.78e+09 & 0.0001 & \textbf{1.1255}\\
     9 & 0.9648 & 2.69e+08 & 0.0507 & 0.7330\\
     3 & 0.9979 & 1.998e+05 & 0.0217 & 0.5327\\
     4 & 0.9890 & 1.18e+06 & 0.0766 & 0.4781\\
    12 & 0.9811 & 6.90e+06 & 0.0586 & 0.3263\\
    \bottomrule
  \end{tabular}
\end{table}

\noindent\textbf{Key insights (FLR on Adult, \(k=6\)).}
\begin{itemize}
  \item \textbf{Exceptional separation quality:} The silhouette score of 0.9443 indicates very tight, well‐separated clusters in the fairness–performance space, suggesting clear trade‐off regimes.
  \item \textbf{Trade‐off extremes:}  
    \begin{itemize}  
      \item \textbf{Cluster 0:} Highest SDP (\(\approx0.40\)) but lowest ROC AUC (\(\approx0.68\)), representing ultra‐fair yet poorly discriminative models.  
      \item \textbf{Cluster 5:} Highest ROC AUC (\(\approx0.80\)) but lowest SDP (\(\approx0.10\)), reflecting maximum accuracy at the expense of fairness.  
    \end{itemize}
  \item \textbf{Mid‐range consistency:} Intermediate clusters show very low internal variance, indicating that moderate fairness–accuracy balances are achieved via consistent feature‐importance patterns under FLR.
\end{itemize}

\FloatBarrier

\subsubsection{Bank Marketing Dataset}
We also applied FLR on \emph{Bank Marketing}, clustering models under age-only fairness.  

\begin{figure}[!htb]
  \centering
  \includegraphics[width=0.6\linewidth]{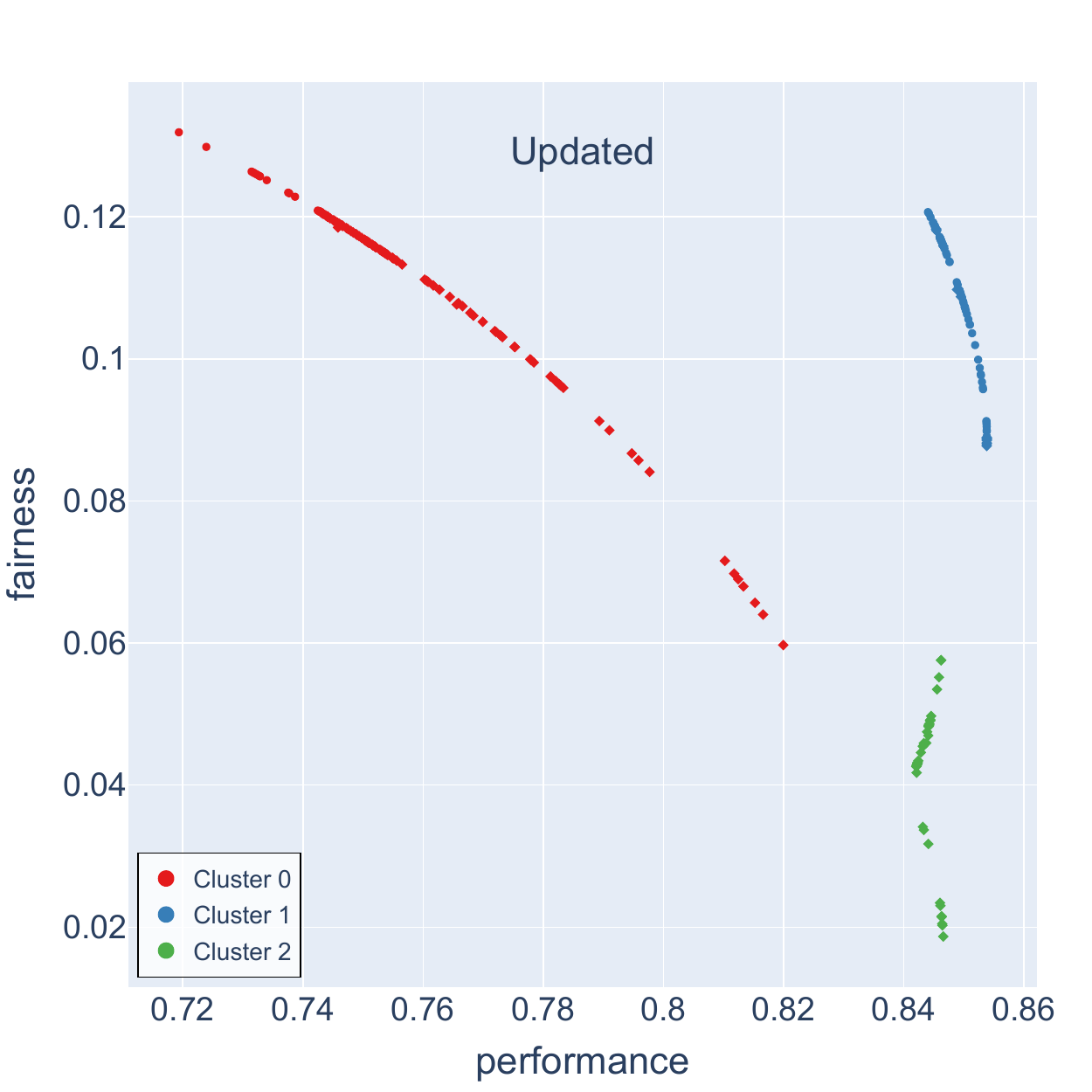}
  \caption{Fair Logistic Regression on the Bank Marketing dataset: clustered model distributions in fairness-performance space for optimal \(k=3\).}
  \label{fig:pareto_flr_bank}
\end{figure}

\begin{table}[!ht]
  \centering
  \scriptsize
  \caption{Composite validation metrics for FLR–Bank Marketing (top-5 by composite).}
  \label{tab:comp_flr_bank}
  \begin{tabular}{rcccc}
    \toprule
    \(k\) & Sil~($\uparrow$) & CH~($\uparrow$) & DB~($\downarrow$) & Comp~($\uparrow$) \\
    \midrule
     3 & 0.8525 & 2874.8  & 0.2869 & \textbf{1.2115}\\
     4 & 0.7908 & 3220.7  & 0.3165 & 0.9068\\
     9 & 0.7063 & 6853.9  & 0.3312 & 0.4086\\
    12 & 0.6747 &10240.0  & 0.3793 & 0.3362\\
    13 & 0.6714 &11821.2  & 0.4066 & 0.2774\\
    \bottomrule
  \end{tabular}
\end{table}

\noindent\textbf{Key insights (FLR on Bank Marketing, \(k=3\)).}
\begin{itemize}
  \item \textbf{Coarse archetype classification:} Only three clusters suffice to capture the FLR trade‐off landscape under age‐only fairness, suggesting more binary regimes in this simpler setting.
  \item \textbf{Clear but less extreme separation:} With a silhouette of 0.8525, clusters are still well‐defined but exhibit slightly more overlap than on Adult, reflecting a narrower fairness span.
  \item \textbf{Archetype definitions:}
    \begin{itemize}
      \item \textbf{Cluster 0:} High fairness (\(\approx0.12\)) with lower accuracy (\(\approx0.72\)), suited for strict parity requirements.
      \item \textbf{Cluster 1:} Balanced middle ground (\(\text{ROC AUC}\approx0.78\), SDP \(\approx0.07\)), ideal for moderate trade‐offs.
      \item \textbf{Cluster 2:} High accuracy (\(\approx0.85\)) with reduced fairness (\(\approx0.02\)), for performance‐driven applications.
    \end{itemize}
  \item \textbf{Evenly spaced trade‐off gaps:} The three clusters align with roughly equal intervals along both axes, making it easy for stakeholders to pick a clear “low,” “medium,” or “high” fairness/accuracy setting.
\end{itemize}

\FloatBarrier

\end{document}